\documentclass{article}

\usepackage{arxiv}

\usepackage[utf8]{inputenc} 
\usepackage[T1]{fontenc}    
\usepackage{hyperref}       
\usepackage{url}            
\usepackage{booktabs}       
\usepackage{amsfonts}       
\usepackage{nicefrac}       
\usepackage{microtype}      
\usepackage{lipsum}
\usepackage{float}
\usepackage{subfig}
\usepackage{fancyhdr}       
\usepackage{graphicx}       
\graphicspath{{media/}}     
\usepackage[utf8]{inputenc}
\usepackage{authblk}
\usepackage{amsmath,amssymb}
\usepackage[normalem]{ulem}
\usepackage[dvipsnames]{xcolor}
\usepackage{graphicx}

\usepackage{caption}
\usepackage[ruled,vlined]{algorithm2e}

\DeclareMathOperator{\EX}{\mathbb{E}}
\DeclareMathOperator*{\argmax}{arg\,max}

\title{Generating GPU Compiler Heuristics using Reinforcement Learning}

\author{
  \textbf{Ian Colbert\thanks{Corresponding author: ian.colbert@amd.com}, ~~Jake Daly, ~~Norm Rubin} \\
  Advanced Micro Devices, Inc. \\
}

\begin{document}
\maketitle

\begin{abstract}
GPU compilers are complex software programs with many optimizations specific to target hardware.
These optimizations are often controlled by heuristics hand-designed by compiler experts using time- and resource-intensive  processes.
In this paper, we developed a GPU compiler autotuning framework that uses off-policy deep reinforcement learning to generate heuristics that improve the frame rates of graphics applications.
Furthermore, we demonstrate the resilience of these learned heuristics to frequent compiler updates by analyzing their stability across a year of code check-ins without retraining.
We show that our machine learning-based compiler autotuning framework matches or surpasses the frame rates for 98\% of graphics benchmarks with an average uplift of 1.6\% up to 15.8\%.
\end{abstract}


\section{Introduction}
\label{sec:introduction}

The benefits of compiler optimizations often depend on characteristics of the code and the capabilities of the target hardware.
For example, an optimization that reorders instructions to increase the number of overlapped loads can often speed up execution.
However, as the number of parallel threads increases, this optimization can exceed hardware limits and slow down execution.
Often, compiler optimizations are controlled by complex decision functions, typically referred to as heuristics, that are hand-crafted by experts.
As the number of code optimizations in production compilers has grown, compiler writers have sought ways to automate the process of tuning these heuristics through autotuning - a technique used to explore code optimizations based on an objective function.

\begin{figure}[H]
	\centering
	\includegraphics[width=\linewidth]{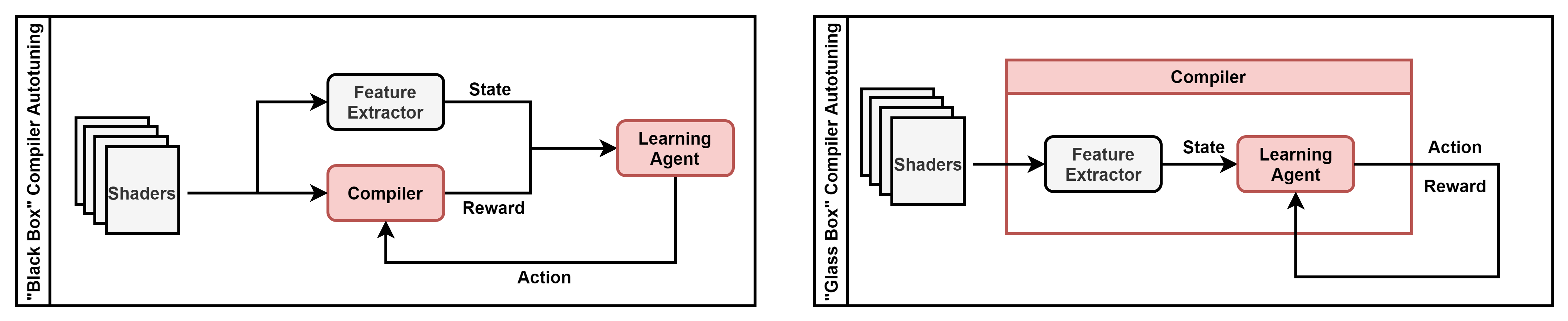}
	\caption{\small{\textbf{Compiler Autotuning Frameworks.} The ``black box" framework injects code optimization flags to externally tune options for a hardware target.
Alternatively, the ``glass box" framework integrates the learning agent in the compiler to act directly on the intermediate representation (IR).}}
	\label{fig:compiler_autotuning_examples}
\end{figure}

{As shown in {Figure~\ref{fig:compiler_autotuning_examples}}, we separate standard autotuning frameworks into "black box" or "glass box" paradigms.
"Black box" compiler autotuning frameworks inject pragmas and code optimization flags to externally tune options to maximize performance for a hardware target.
In this paradigm, the learning agent (\textit{e.g.}, a human expert or computational model) is external to the compiler.
However, this limits the benefits of a well-tuned solution when new code is introduced in production as the learning agent is often not shipped with the compiler.
Alternatively, "glass box" compiler autotuning frameworks integrate the learning agent in the compiler, where a well-trained solution can act on new code introduced when deployed in a production setting.}

As such, we have focused on two problems under this "glass box" paradigm: (1) learning stable GPU compiler heuristics through reinforcement learning-based (RL-based) autotuning and (2) deploying these trained models as heuristics in production compilers.
To our knowledge, this is the first ML-based autotuning framework used to learn, integrate, and deploy heuristics in a production GPU compiler.
Below, we summarize our contributions:

\begin{enumerate}
\item We formulated production GPU compiler autotuning as an off-policy reinforcement learning problem to maximize the reuse of performance data in a production setting (Section~\ref{sec:formulation}).

\item We designed an autotuning framework that addresses both the unique complexities that arise when optimizing for multi-threaded hardware as well as the stability required of deployed learned heuristics in constantly evolving production compilers (Section~\ref{sec:framework}).

\item We applied our framework to a unique compiler optimization over two GPUs and deployed the resulting models in the production AMD Radeon\texttrademark~Software graphics compiler (Section~\ref{sec:experimental-results}).

\item We show the inherent resilience of our trained models to software updates by analyzing the stability of our deployed compilers heuristics over a year of production code check-ins without retraining (Section~\ref{sec:experimental-results}).

\end{enumerate}

\section{Background}
\label{sec:context}

One could be guaranteed to find the optimal heuristic settings by exhaustively searching the code optimization space using iterative trial-and-error.
However, the cost of these experiments quickly becomes a limiting factor as graphics benchmarks have non-trivial runtimes and production compilers are constantly evolving.
In practice, standard approaches to compiler heuristic autotuning follow expert-driven or machine learning-based (ML-based) strategies.
Using expert-driven autotuning, compiler optimizations are controlled by heuristics typically hand-tuned over a focused subset of representative benchmarks that are selected and analyzed by compiler (or hardware) experts.
This approach requires a significant amount of expertise as optimizations become more complex and may fail to exploit complex high-dimensional patterns.
Alternatively, machine learning offers a robust means of automatically learning non-linear decision functions in high dimensional spaces.
This makes it an ideal fit for compiler autotuning problems, where the space to tune a single compiler optimization heuristic is {non-linear} with many local minima and can vary across target hardware~\cite{bodin1998iterative}.
Given a representative and accurately labeled dataset, ML-based autotuning can generalize to new programs easier and be re-tuned for new hardware faster~\cite{leather2020machine}.
In a "glass box" compiler autotuning framework, these learned models are later integrated into the compiler to control code optimizations as heuristics.

\subsection{Production GPU Compiler Autotuning}
\label{sec:background_compiler_autotuning}

Modern graphics applications such as computer games {contain} dedicated kernels, referred to as shaders, that concurrently execute as pipelined functions when rendering a frame.
The frames rendered by these applications represent views of a virtual 3D scene defined by complex geometry, material properties, and light sources.
The shader pipelines used to render these frames are designed to project the geometric primitives of the 3D virtual scene (\textit{e.g.}, points, lines, and triangles) onto the 2D screen and calculate the color and opacity of each pixel.
These shaders are written in a high-level device-independent language which makes use of a large, filtered look-up table parameterized by a collection of multi-dimensional arrays referred to as resources.
Often, before the application ships, these shaders are compiled by a front-end compiler into device-independent byte code.
Later, these byte codes are separately used as inputs to a supplied driver which then translates them into machine-dependent byte codes to finally be compiled by the back-end compiler into a target-specific instruction set architecture (ISA).
The ISA and dynamic resource information is then packaged together and handed off to the GPU for execution.
Additionally, the application can execute one or more pipelines asynchronously.

Recent work has tried to capture dynamic information to guide program optimization~\cite{stephenson2021cpgo}; however, neither resource information nor concurrently executing shader pipelines are known to the back-end compiler at compile time.
Consequently, heuristics controlling code optimizations are simplified by assuming that speeding up each program individually leads to speed-ups on the application globally.
{While this is often true for single-threaded applications}, {GPU graphics applications are highly parallel} multi-threaded optimization problems where programs execute concurrently and compete for shared hardware resources.
In this environment, focusing on improving the execution time of an individual program in isolation will not always lead to speed-ups on the application globally.
Allocating resources either detracts from the shared pool or reduces the ability for multiple programs to run concurrently.

In this paper, we view the back-end compiler and driver as a single object.
As such, we refer to both the back-end compiler and the driver as "the compiler".
The non-deterministic nature of graphics applications, the multi-threaded execution model of the GPU, and the constant evolution of production software introduce unique complexities on top of standard compiler autotuning problems.
We summarize these complexities as follows:

\begin{enumerate}
	\item \textbf{Decisions made by the front-end compiler cannot be undone.} Different front-end compilers can generate radically different byte code, all of which have to be supported and optimized by the back-end compiler.
	
	\item \textbf{There are many opportunities for parallel processing that are difficult to account for.} The unpredictable cost of executing each shader and the constraints on the order of their execution introduce dynamic dependencies that complicate parallel implementations.

	\item  \textbf{The computational and bandwidth requirements can vary significantly depending on the behavior of shaders and properties of a scene.} The compiler has very little insight into this; for example, if the same shader is used in multiple pipelines, the compiler typically cannot optimize them differently in each pipeline.
	
	\item \textbf{The frequency of production compiler updates can change and inadvertently destabilize tuned heuristics.} As production compilers are updated, their intermediate representation (IR) of each shader can change.
As such, heuristics deployed in production compilers need to be robust to these shifts in compiler IR.
\end{enumerate}

We address this problem of per-shader code optimization under the unique conditions that arise from production GPU compiler autotuning.

\subsection{Supervised Learning}
\label{subsec:supervised-learning}

The majority of ML-based autotuning frameworks focus on supervised learning (SL) approaches to fit a predictive model to labeled data.
This strategy has provided state-of-the-art performance in the field of compiler autotuning~\cite{ashouri2018survey, ashouri2014bayesian, bergstra2012machine, cavazos2007rapidly, cummins2017end, fursin2011milepost, leather2014automatic, monsifrot2002machine}.
Under the SL training paradigm, inputs to the model are derived from characteristics of program code and/or the target hardware while labels are derived from performance measurements.
However, the success of SL on discriminative tasks is reliant on the quality of the training dataset~\cite{song2020learning,zhang2016understanding,zhu2004class}.
As shown by Zhang \textit{et al.}~\cite{zhang2016understanding}, DNNs can fit to entire training sets even in the presence of corruption, which leads to poor generalizability when deployed in the real world.
In the scope of GPU compiler autotuning for real-world graphics applications, accurate labels derived from frame rates or execution times are either unknown or become computationally intractable to determine with the non-deterministic multi-threaded execution model.

\subsection{Reinforcement Learning}
\label{subsec:reinforcement-learning}

\begin{figure}[H]
\begin{center}
\includegraphics[width=0.5\linewidth]{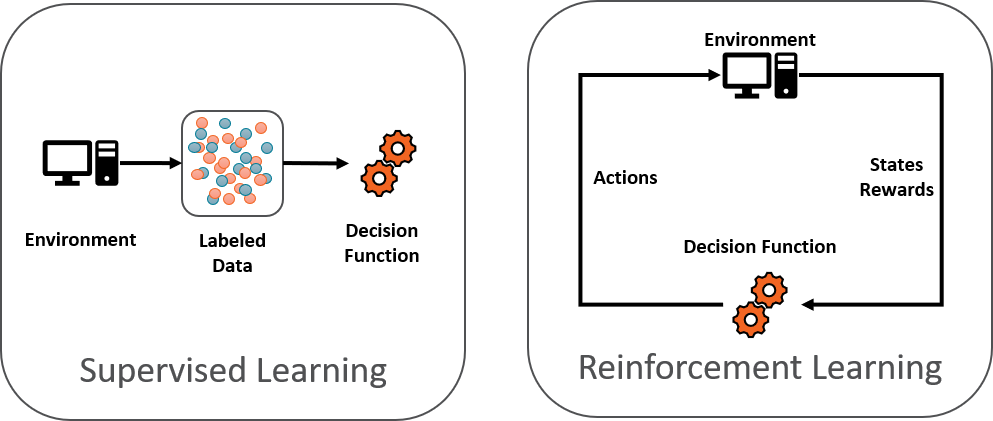}
\caption{\textbf{Supervised Learning (SL) vs. Reinforcement Learning (RL).} Under a SL paradigm, a predictive model is fit to labeled data previously collected to represent a target environment. Under a RL paradigm, a decision function is trained to select the action for a given state that maximizes the expected reward through trial-and-error interactions with a target environment.}
\label{fig:learning-strategies}
\end{center}
\end{figure}

Unlike SL, reinforcement learning (RL) algorithms do not rely on accurate pre-determined labels, but rather on a reward signal received from interacting with an environment by observing states and selecting actions in accordance to a decision function, as shown in Figure~\ref{fig:learning-strategies}.
The decision function, referred to as a policy, is used to map states to actions and, in deep reinforcement learning (DRL), can be modeled using a DNN.
The DRL training process aims to optimize an objective function by trial-and-error using reward signals generated by the environment to guide parameter updates~\cite{graesser2019foundations}.
The objective under this paradigm is to learn a policy ($\pi$) that maximizes the cumulative expected rewards for a sequence of states, typically referred to as a trajectory ($\tau$).
The "state, action, reward" tuples are generated through a controlled feedback loop between the policy and the environment.

RL algorithms are typically applied to sequential decision-making problems but have been shown to achieve high performance in single-time step domains such as classification or detection tasks~\cite{lin2020deep,zhao2016deep}.
Unlike SL algorithms which train on large static datasets, RL algorithms often dynamically collect data throughout training\footnote{It is important to note that there are cases in which supervised learning algorithms train on dynamically changing datasets - online learning strategies, for example. There are also applications where reinforcement learning learn entirely on static datasets.}.
As discussed further in Section~\ref{sec:related-works}, select previous works approach compiler autotuning using DRL~\cite{haj2020neurovectorizer,huang2019autophase,mcgovern2002building,trofin2021mlgo}.
Here, states are derived from characteristics of program code, actions are applied code optimizations or heuristic settings, and rewards are derived from performance measurements.

\subsubsection{Q-Learning}
\label{sec:q-learning}

$Q$-Learning is a simple, computationally efficient RL strategy that learns how to act optimally by iteratively improving evaluations of the quality of particular actions at particular states, often referred to as state-action pairs~\cite{watkins1992q}.
In discrete action spaces, these evaluations are stored in a data structure often referred to as a $Q$-table\footnote{In domains with continuous action spaces, this is often referred to as a $Q$-function and modeled using a DNN~\cite{mnih2013playing}}, where the value of each state-action pair is calculated using the expected cumulative reward discounted by a time factor ($\gamma$) when taking action $a$ in state $s$ using policy $\pi$, as shown in Eq.~\ref{eq:q-function-def}.
Here, each element is interpreted as the estimated value of a given action at a given state when following policy $\pi$ through trajectory $\tau$.
When greedily following the $Q$-table, an agent chooses the sequence of actions that maximize the expected reward from initial state $s_0$.
Watkins \textit{et al.}~\cite{watkins1992q} prove in detail that, in discrete action spaces,  $Q$-learning converges to the optimal values of state-action pairs with probability 1 given that all state-action pairs are repeatedly sampled.
These optimal values constitute the optimal $Q$-table, often referred to as $Q^*(s,a)$, and are interpreted as the expected cumulative discounted reward when following the optimal policy $\pi^*$ through a trajectory $\tau$ of size $T$.

\begin{equation} \label{eq:q-function-def}
Q^{\pi} (s,a) = \EX_{s_0=s,a_0=a}[\sum_{t \in \tau} \gamma^t r_t]
\end{equation}

\section{Problem Formulation and Solution Overview}
\label{sec:formulation}

Standard DRL algorithms are \textit{active} learning processes which requires frequent interaction with a stationary environment~\cite{kumar2020conservative}.
Throughout the trial-and-error training process, the data collected when interacting with the environment using the current policy can only be used to update internal parameters once~\cite{graesser2019foundations}.
As such, when the policy changes (or the environment changes), the data can no longer be used.
By continually discarding old data, this class of learning algorithms, referred to as "on-policy" DRL, is sample inefficient~\cite{graesser2019foundations}.
Gleeson \textit{et. al}~\cite{gleeson2021rl} show that on-policy algorithms are at least 3.5x more simulation-bound than off-policy algorithms.
If we were to use on-policy DRL for GPU compiler autotuning, the system would be heavily bottlenecked by the duration of graphics benchmarks, which can each run several minutes.
With the frequency of production compiler updates and as the number of code optimizations under consideration increases, the detrimental impact of this bottleneck exponentially increases.

Alternatively, "off-policy" DRL algorithms learn from previously collected data without reliance on frequent interactions, even with updated policies in non-stationary environments.
For this reason, we focus on off-policy DRL for GPU compiler autotuning and separate data collection and model training to inexpensively leverage pre-existing performance automation workflows to build offline datasets to train on.
However, without corrective feedback, off-policy reinforcement learning strategies are notoriously instable and sensitive to hyperparameter tuning~\cite{kumar2020conservative,kumar2020discor}.
To both reintroduce corrective feedback and alleviate data collection bottlenecks, we introduce the $Q$-learning strategy discussed in Section~\ref{sec:framework} that decouples data collection and model training to each run continuously in a controlled feedback loop.

\subsection{Problem Formulation}

{The goals of our RL-based GPU compiler autotuning training strategy are: (1) to determine the heuristic setting for a given shader that yields the highest average frame rate increase over a target set of graphics applications; and (2) to deploy the learned decision function as a compiler heuristic that is stable in the presence of frequent code changes.}
To this end, we define the following: \\

\noindent \textbf{Environment:} Unlike standard DRL problems, which assume a stationary environment, frequent code changes create a dynamic environment that is non-stationary, although the target hardware is static.
As such, we refer to the target hardware and the dynamically changing production compiler as the non-stationary environment $\mathcal{E}$.
We refer to a compiler built at revision $t$ as $\mathcal{E}_t$ such that $\mathcal{E}_t \in \mathcal{E}$, where $t$ is a time step measured in code check-ins.

\noindent \textbf{States:} We derive the state ($s$) from the compiler intermediate representation (IR) of a shader as further described in Section~\ref{sec:data_collection}.
In production, as the compiler is updated, the same shader may be compiled into different IR and, therefore, many possible states.

\noindent \textbf{Action:} We define the action ($a$) as the heuristic setting applied to the IR during compilation.
We define the actions taken when greedily following the optimal policy ($\pi^*$) as the \textit{optimal} action ($a^*$), further described in Section~\ref{sec:training}.

\noindent \textbf{Rewards:} We derive the reward ($r$) from the observed frame rate when taking action $a$ for state $s$, further discussed in Section~\ref{sec:rewards}.

\noindent \textbf{Trajectory:} Because the compilation strategy of modern GPU compilers addresses each shader program in isolation, we define each trajectory ($\tau$) as a single time-step Markov decision process (MDP) such that $\tau_i = \{s_i\}$.

\noindent \textbf{Policy:} We refer to the decision function that maximizes the expected frame rate improvement over a given set of graphics applications ($\mathcal{A}$) executed in environment $\mathcal{E}_t$ as the \textit{optimal} policy ($\pi^*$). 
Our decision policy ($\pi_\theta$) throughout training is modeled using a DNN with learnable parameters $\theta$.
We interpret this decision policy as the probability of taking action $a$ for a given state $s$ such that $\pi_\theta\left(a| s\right)$.

\subsection{Solution Overview}

\begin{figure*}
\centering
\includegraphics[width=0.9\linewidth]{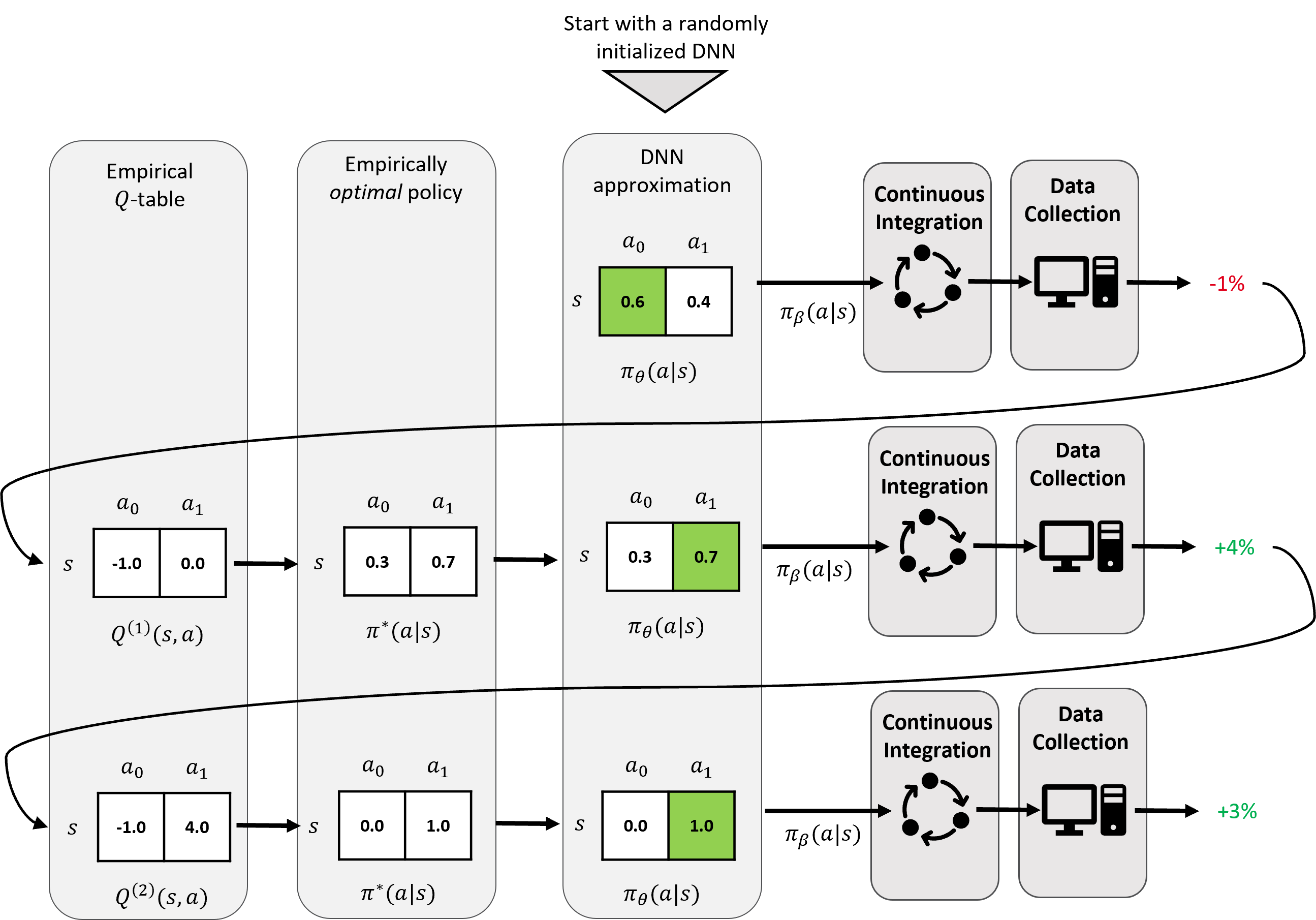}
\caption{\textbf{RL-based GPU Compiler Autotuning Framework.} Our RL-based compiler autotuning framework combines deep reinforcement learning (DRL) with continuous integration (CI) to deploy trained models as heuristics in production compilers.}
\label{fig:autotuning-framework}
\end{figure*}

{We introduce the training strategy depicted in Figure~\ref{fig:autotuning-framework} and detailed in Section~\ref{sec:training} to generate GPU compiler heuristics using an off-policy reinforcement learning (RL) algorithm based on $Q$-learning.
The training objective under this paradigm is to find the optimal $Q$-table that maximizes the expected frame rate improvement when applying heuristic setting $a$ to a given state $s$.
The inference objective is to then fit a DNN decision policy ($\pi_\theta$) to the empirically estimated optimal policy ($\pi^*$) by learning the set of parameters $\theta$ that minimize its divergence from $\pi^*$.
Once trained offline, the learned inference model is then integrated directly into the compiler to act as a heuristic decision function. 
For clarity, we differentiate between decision policy $\pi_\theta\left(a|s\right)$ and behavior policy $\pi_\beta\left(a|s\right)$.
We define the behavior policy $\pi_\beta$ as the inference model counterpart to $\pi_\theta$, where the parameters of $\pi_\beta$ are pre-trained and frozen from the decision policy.
Whereas the decision policy is iteratively updated throughout training, the behavior policy is frozen at inference time.}

\section{RL-based GPU Compiler Autotuning}
\label{sec:framework}

The non-trivial cost of running a GPU graphics benchmark requires an autotuning framework that is both robust to an evolving compiler and generalizable across an application suite.
To learn stable heuristics, our RL-based GPU compiler autotuning framework, given by Algorithm~\ref{alg:heuristic-optimization} and depicted in Figure~\ref{fig:autotuning-framework}, consists of three modules pipelined in a controlled feedback loop - continuous integration, data collection, and model training.
Starting from a randomly initialized decision policy ($\pi_\theta$), we use continuous integration (CI) to deploy the behavior policy ($\pi_\beta$) as a heuristic in the latest production compiler (see Section~\ref{subsec:continuous-integration}).
We then collect the observed performance measurements and intermediate representations (IR) of the shaders in each graphics benchmark in our set of applications ($\mathcal{A}$) (see Section~\ref{sec:data_collection}).
After updating (or expanding) the $Q$-table, we train $\pi_\theta$ to approximate the empirically optimal decision policy ($\pi^*$) to maximize the expected frame rate over $\mathcal{A}$ (see Section~\ref{sec:training}).
The learned parameters of $\pi_\theta$ are periodically used to update $\pi_\beta$ to be deployed as a heuristic in the latest compiler using CI to restart the feedback loop.

\begin{algorithm}
\SetAlgoLined
\KwResult{$\pi_\beta^{(N)}$}
 Randomly initialize $\pi_\theta$\;
 Initialize an empty $Q$-table\;
 \While{$i < numIterations$}{
  Copy learned parameters of $\pi_\theta$ to $\pi_\beta^{(i)}$  (Section~\ref{subsec:continuous-integration})\;
  Integrate $\pi_\beta$ into the latest compiler (Section~\ref{sec:inference})\;
  Get states and performance measurements over $\mathcal{A}$ (Section~\ref{sec:data_collection})\;
  Update (or expand) the $Q$-table based on the observed performance measurements (Section~\ref{sec:q_learning_autotuning})\;
  Convert the empirical $Q$-table to probabilities $\pi^*(s, a)$ (Section~\ref{sec:q_learning_autotuning})\;
  Train $\pi_\theta(a|s)$ to approximate $\pi^*(s, a)$ (Section~\ref{sec:approx})\;
  }
 \caption{RL-based GPU Compiler Autotuning}
 \label{alg:heuristic-optimization}
\end{algorithm}

\subsection{Continuous Integration}
\label{subsec:continuous-integration}

Continuous integration (CI) is the large-scale software engineering practice of merging developers' working source code revisions into a shared mainline.
Standard practice adopts a development strategy of merging these changes several times a day.
As such, the development of production-level compilers is fast paced, which can lead to significant changes in resulting code as discussed further in Section~\ref{sec:production_compilers}.
Standard ML-based autotuning efforts fit the decision function to a frozen snapshot in time.
However, the possibility of freezing a production compiler is precluded by the team of engineers dedicated to improving it through continuous updates.
In Section~\ref{sec:q_learning_autotuning}, we discuss how we account for CI throughout training.

\subsubsection{Inference Engine}
\label{sec:inference}

\begin{figure}
\begin{center}
\includegraphics[width=0.6\linewidth]{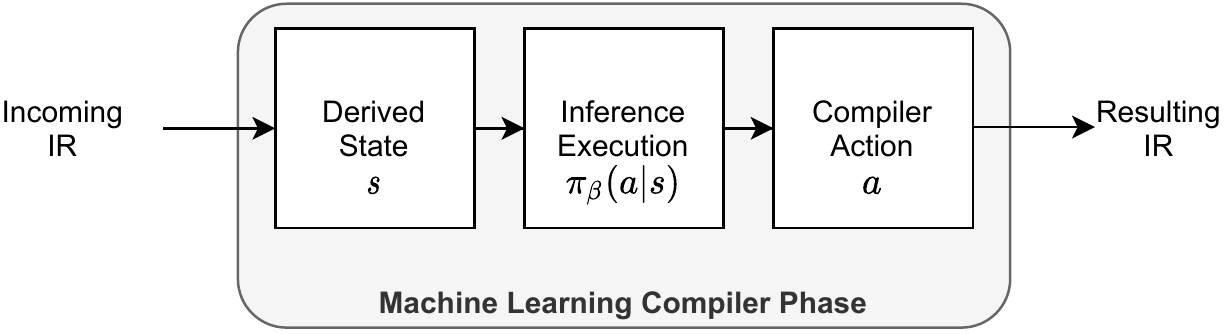}
\caption{{\textbf{Machine Learning Compiler Phase.} The behavior policy ($\pi_\beta$) is deployed as a compiler heuristic to be applied directly to the IR. The state ($s$) is derived directly from the incoming IR to be analyzed by $\pi_\beta$. The model output is then used to determine the compiler action ($a$).}}
\label{fig:scml-phase}
\end{center}
\end{figure}

Figure~\ref{fig:scml-phase} depicts the compiler inference engine, where the behavior policy ($\pi_\beta$) is deployed as a compiler heuristic to be applied directly to the IR, before any machine-dependent optimizations.
The state is derived from the incoming IR (see Section~\ref{sec:data_collection}) to be analyzed by $\pi_\beta$ where the model output is then used to determine which heuristic setting can be optimally applied.
Using continuous integration (CI) ensures $\pi_\beta$ is always integrated into the latest compiler for each performance run.
As shown in Figure~\ref{fig:autotuning-framework}, each trial-and-error loop collects the derived static features and the applied compiler heuristic settings ($a$) for every shader as well as the observed frame rates ($r$) of every graphics application in our benchmark suite ($\mathcal{A}$).
These static features are derived from the incoming IR before the neural network forward pass and used to build each state ($s$).

\subsubsection{Robust Reward Signals}
\label{sec:rewards}

The success of learning a stable policy that generalizes well to real-world data is dependent on an informative reward signal that reinforces desired behavior.
Under a multi-threaded execution model, speeding up one shader can sometimes lead to significant performance degradation in another when co-executing programs compete for shared resources.
Because our objective is to maximize the frame rate of graphics applications rather than reducing shader execution time, we define our reward signal ($r$) as the normalized change in frame rate observed when compared to the global default compiler actions.
Given a baseline frame rate of $F_0$ and an observed frame rate of $F$, the reward function becomes the relative speed-up given by $\frac{F}{F_0}$.
However, as compilers evolve, previously collected performance measurements become stale.
We further discuss this in Section~\ref{sec:q_learning_autotuning}.

\subsection{Data Collection}
\label{sec:data_collection}

Approaches to representing source code for machine learning models can be split into static and dynamic techniques.
Dynamic techniques rely on performance counters which offer a condensed summary of the program’s behavior at runtime and have been empirically shown to outperform methods relying strictly on static features alone~\cite{ashouri2014bayesian,cavazos2007rapidly}.
However, dynamic performance counter information is expensive to collect and unrealistic in production compilers that do not have that information available at compile time.
Alternatively, static techniques aim to form a representation directly from the source code itself.
Recent works have drawn from the field of natural language processing to learn distributed representations of instructions for automated feature extraction~\cite{cummins2017end,leather2014automatic,alon2019code2vec,ben2018neural,cummins2020programl, venkatakeerthy2020ir2vec}.
While effective, these approaches can consume large chunks of memory and require more compute resources than are available in practice.
To operate within the compute and memory constraints of production compilers, we capture fast, machine-independent features such as total instructions, number of basic blocks, and memory operations directly from the compiler IR at compile time.
By doing so, we tightly represent the compiler IR of each shader as a fixed-length feature vector at compile time using only 44 input features.
When running the model at 32-bit floating point precision, this only consumes 176 bytes.
Table~\ref{tbl:feature-list} provides a high-level summary of the categories of attributes used, each of which can be determined independently of target hardware.

\begin{table}
\begin{center}
\begin{tabular}{ |l| }
 \hline
 \# of basic blocks \\
 \# of vector instructions \\
 \# of scalar instructions \\
 \# of memory instructions \\
 \# of compute instructions \\
 \# of control flow instructions \\
 \# of registers used \\
 \# of work groups \\
 The shader hardware stage \\
 \hline
\end{tabular}
\end{center}
\caption{\textbf{Summary of Static Feature Categories Considered.} Here, we provide a summary of the categories of static features considered when representing the incoming IR as a fixed-length feature vector.}
\label{tbl:feature-list}
\end{table}

\subsection{Model Training}
\label{sec:training}

Unlike on-policy DRL algorithms, which require $\pi_\theta = \pi_\beta$, off-policy algorithms can learn from previously collected data independent of the behavior policy $\pi_\beta$ such that $\pi_\theta \neq \pi_\beta$.
This allows data to be collected asynchronously and re-used throughout training iterations.
In production systems, this is important since data collection does not bottleneck training and a decision policy can learn over historical revisions.
However, as further discussed in Section~\ref{subsec:continuous-integration}, even mature production compilers are constantly changing.
To deploy stable heuristics robust to changes in the compiler IR, we implement a DRL strategy based on $Q$-learning.

\subsubsection{Q-Learning for GPU Compiler Autotuning}
\label{sec:q_learning_autotuning}

{Here, we formalize $Q$-tables for RL-based GPU compiler autotuning.
Let $P(a|s; \mathcal{A})$ denote the probability that applying a heuristic setting ($a$) to a shader given its state ($s$) will maximize the expected frame rate over a set of applications ($\mathcal{A}$).}
The optimal heuristic setting ($a^*$) is then the action that maximizes this conditional probability, as shown in Eq.~\ref{eq:best-heuristic-setting}.
Given that we have defined each trajectory as a single time-step MDP, the $Q$-table (see Eq.~\ref{eq:q-function-def}) under this formulation is independent of future sequences of states and actions.
{As shown in Eq.~\ref{eq:q-function-t0-mdp}, the resulting value of each state-action pair can be simplified to its expected reward over $\mathcal{A}$.}
{The optimal decision policy $\pi^*(s, a)$ can then greedily follow this $Q$-table to maximize the expected reward.}
Alternatively, the joint probability distribution $\pi^*(s, a)$ can be derived from $Q(s,a)$ using the standard Boltzmann softmax operator, as given by Eq.~\ref{eq:action-prob-mm}, where $\rho$ is a temperature parameter such that $\rho \in (0,\infty)$.
As $\rho$ increases, the derived policy moves towards a uniform distribution.
As $\rho$ decreases, it moves closer towards the greedy policy given by Eq.~\ref{eq:best-heuristic-setting}.

\begin{equation}
\label{eq:best-heuristic-setting}
a^* = \argmax_{a} ~P(a|s;\mathcal{A})
\end{equation}

\begin{equation}
Q(s,a) = \EX[ r | s, a; \mathcal{A} ]
\label{eq:q-function-t0-mdp}
\end{equation}

\begin{equation} \label{eq:action-prob-mm}
\pi^*(a|s) = \frac{e^{Q(s,a)/\rho}}{\sum_{a_j} e^{Q(s,a_j)/\rho}}
\end{equation}

As the compiler updates, performance measurements become stale.
To account for this throughout training, we apply a discount ($\omega$) based on the change in time ($\Delta t$) as measured in code check-ins.
As data is collected from the $n^\text{th}$ iteration, the $Q$-table values are updated using Eq.~\ref{eq:table-update} if the state already exists in the data structure such that $s \equiv s_n$ and $a \equiv a_n$.
\begin{equation} \label{eq:table-update}
    Q_{n}(s, a) = (1 - \alpha) \cdot \omega^{\Delta t} \cdot Q_{n-1}(s, a) +  \alpha \cdot r_n
\end{equation}

If the state has yet to be seen, it is initialized to the observed reward ($r_n$).
Here, $\alpha$ is learning rate and $\Delta t$ is dependent on the rate of code check-ins over a given measurement of time.
Note that $\alpha$ and $\omega$ are tuneable hyperparameters.

\subsubsection{Approximating the Optimal Policy}
\label{sec:approx}

Compiler updates lead to changes in the IR which creates a non-stationary environment ($\mathcal{E}$).
Because the shader state ($s$) is derived from the compiler IR, these changes produce more states, creating a dynamic $Q$-table in response to the non-stationary $\mathcal{E}$.
As the compiler is updated, this $Q$-table would be too big to use as it expands with new state-action pairs.
Additionally, a $Q$-table does not easily generalize to new states since it is a finite-state look-up table.
As shown in Algorithm~\ref{alg:heuristic-optimization} and depicted in Figure~\ref{fig:autotuning-framework}, we train $\pi_\theta$ to approximate the empirically optimal decision policy $\pi^*(s, a)$ by minimizing the Kullback-Leibler divergence as given in Eq.~\ref{eq:loss-function}.
Note that $\pi^*(s, a)$ is derived from the empirically estimated $Q$-table using Eq.~\ref{eq:action-prob-mm}, where the temperature parameter $\rho$ can be decreased throughout training as the table converges to $Q(s,a)$.
The learned parameters of $\pi_\theta$ are periodically deployed as the behavior policy $\pi_\beta$ to collect performance measurements and update the empirical $Q$-table using the DRL trial-and-error feedback loop.

\begin{equation} \label{eq:loss-function}
\mathcal{L} = D_{KL}\left(\pi_\theta || \pi^*\right)
\end{equation}

\begin{figure*}
    \centering
    \includegraphics[width=\linewidth]{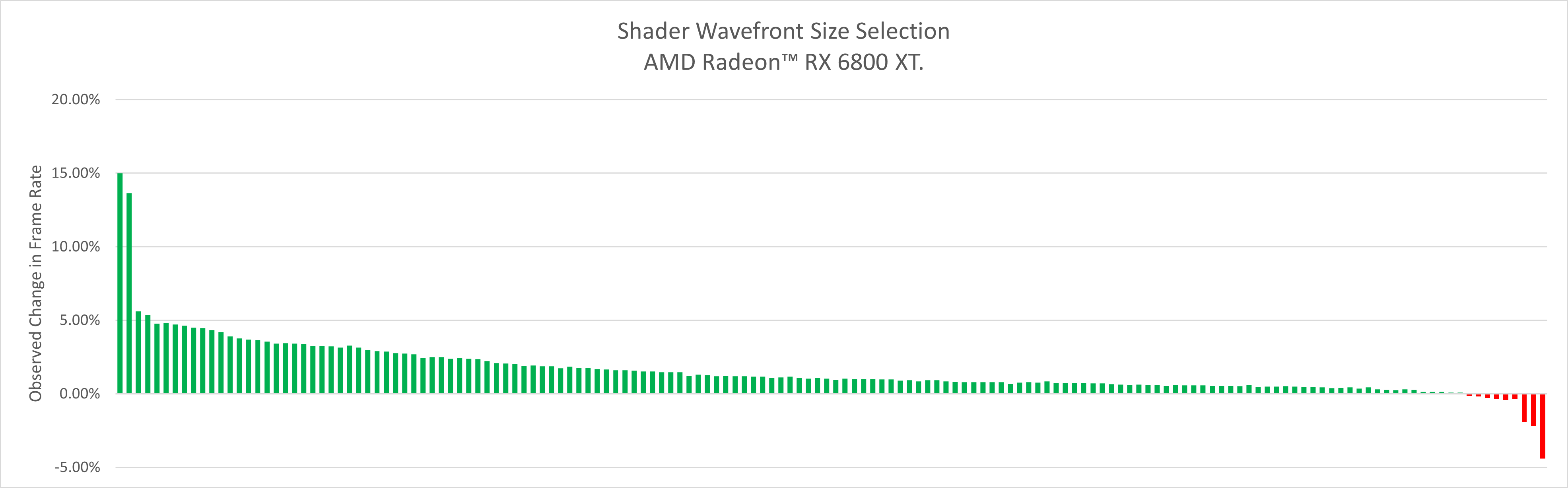}
    \caption{\small{\textbf{Shader Wavefront Size Selection - AMD Radeon\texttrademark~6800 XT.} We provide a histogram of the observed change in frame rate when enabling the RL-based compiler heuristic to select shader wavefront size within the production AMD Radeon\texttrademark~Software graphics compiler across the suite of graphics applications executed on the AMD Radeon\texttrademark~6800 XT.}}
    \label{fig:wavefront_size_selection}
\end{figure*}

\section{Experimental Results}
\label{sec:experimental-results}

Performance gains seen on graphics applications from production compilers are typically the combination of smaller gains aggregated over several tuned heuristic optimizations.
To realize these performance gains, expert-driven autotuning efforts often greedily focus resources on highly impactful shaders.
However, this approach tends to only account for the top percentage of shaders in a graphics benchmark.
Our framework for RL-based autotuning provides a means of automatically exploring more of the optimization space across all shaders in a benchmark application.
We applied our RL-based autotuning framework to a compiler optimization and two GPUs to learn, integrate, and deploy stable heuristics in the production AMD Radeon\texttrademark~Software graphics compiler.

\subsection{Adapting to Production Compiler Development}
\label{sec:production_compilers}

\begin{figure}
\centering
\includegraphics[width=0.4\linewidth]{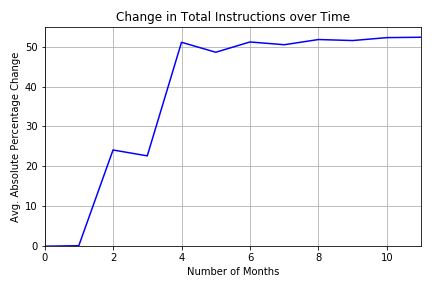}
\caption{\small{\textbf{Average Change in IR Counters over Time.} Over time, code changes to the compiler significantly impact the state space of the of the shader IR. Even the total number of IR instructions in mature production compilers can vary as much as 50\% over the course of a year.}}
\label{fig:change_in_counters}
\end{figure}

\begin{figure}
\begin{center}
\includegraphics[width=0.5\linewidth]{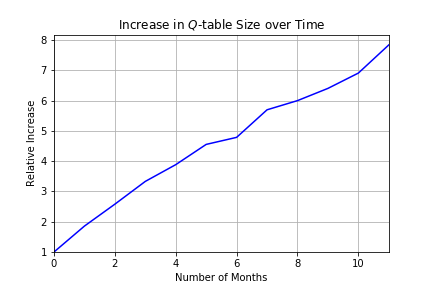}
\caption{\small{\textbf{Relative Increase in $Q$-table Size.} The rapid pace of software development significantly impacts the state space over time. Updates to the production compiler throughout the DRL trial-and-error loop can cause the $Q$-table to grow when new states are derived from changes to compiler IR. Because the $Q$-table requires so much memory, we train $\pi_\theta$ to approximate the optimal policy $\pi^*(s, a)$ to be periodically deployed as $\pi_\beta$ (see Section~\ref{subsec:continuous-integration}).}}
\label{fig:change_in_q_table}
\end{center}
\end{figure}

{The development of production-level compilers can lead to significant changes in generated code, resulting in a dynamic, non-stationary environment.
As described in Section~\ref{sec:formulation}, we define our non-stationary environment as the target hardware and continuously updated production compiler.
As shown Figure~\ref{fig:change_in_counters}, the absolute percentage change in IR instructions for a given shader can vary as much as 50\% over the course of a year's worth of compiler updates.}
This frequent change in IR results in a shift in the derived state ($s$) which incrementally increases the size of the $Q$-table.
{Figure~\ref{fig:change_in_q_table} shows how the $Q$-table grows over a year's worth of compiler updates.}
To avoid this overhead, we approximate the optimal policy ($\pi^*$) with a DNN decision policy ($\pi_\theta$) using the techniques discussed in Section~\ref{sec:approx}.

\subsection{Shader Wavefront Size Selection}
\label{subsec:experiments-main}

The AMD RDNA\texttrademark~graphics architecture supports wavefronts of either 32 or 64 work-items.
The benefits of these two execution modes, which are intuitively referred to as wave32 and wave64, are dependent on dynamic properties of the shader such as divergence and memory access order.
Running shaders concurrently further complicates optimizing a decision function.
Taking bandwidth as an example, a wider wavefront can result in better performance if there is sufficient bandwidth to support memory accesses without cache misses.
Narrower wavefronts can reduce the strain on bandwidth as the two memory accesses can be separated in time.
However, system bandwidth is a dynamic resource consumed by shaders running concurrently.
Thus, the optimal wavefront size choice is dependent on the dynamic environment of shaders executing concurrently throughout the course of a graphics application.

A compiler has no awareness of this dynamic information at compile time and is forced to make decisions off of static information alone.
We applied our RL-based autotuning framework to improve the frame rate over a set of graphics applications by accurately selecting the optimal wavefront size for each shader at compile time.
As described in Section~\ref{sec:data_collection}, we represent shader state ($s$) as a fixed-length vector of static features derived from the compiler IR.
For wavefront size selection, we constrain $a$ to a binary action space such that $a\in\{\text{wave}32,\text{wave}64\}$.
As described in Section~\ref{sec:rewards}, the reward signal $r$ is derived from the observed change in frame rate with respect to the default behavior.
The decision policy $\pi_\theta$ is a resource-efficient, 3-layer feed-forward classifier with less than 20 KB of learnable parameters.
As discussed in Section~\ref{sec:formulation}, the learned parameters are periodically integrated into the compiler as the behavior policy ($\pi_\beta$).

Applying our RL-based GPU compiler autotuning framework to optimizing shader wavefront size selection for AMD's Radeon\texttrademark~6800 XT, the learned compiler heuristic matches or surpasses the frame rates in 98\% of graphics benchmarks with increases of up to 15.8\% and an average of 1.6\%.
The model converged in only 45 iterations through each of the benchmarks in the application suite.
All experiments are run in a production AMD GPU graphics compiler using a benchmark suite of over {150} graphics benchmarks with an average of 230 unique shaders per benchmark.
Figure~\ref{fig:wavefront_size_selection} gives a histogram of the results.

\begin{figure*}[t!]
    \centering
    \includegraphics[width=\linewidth]{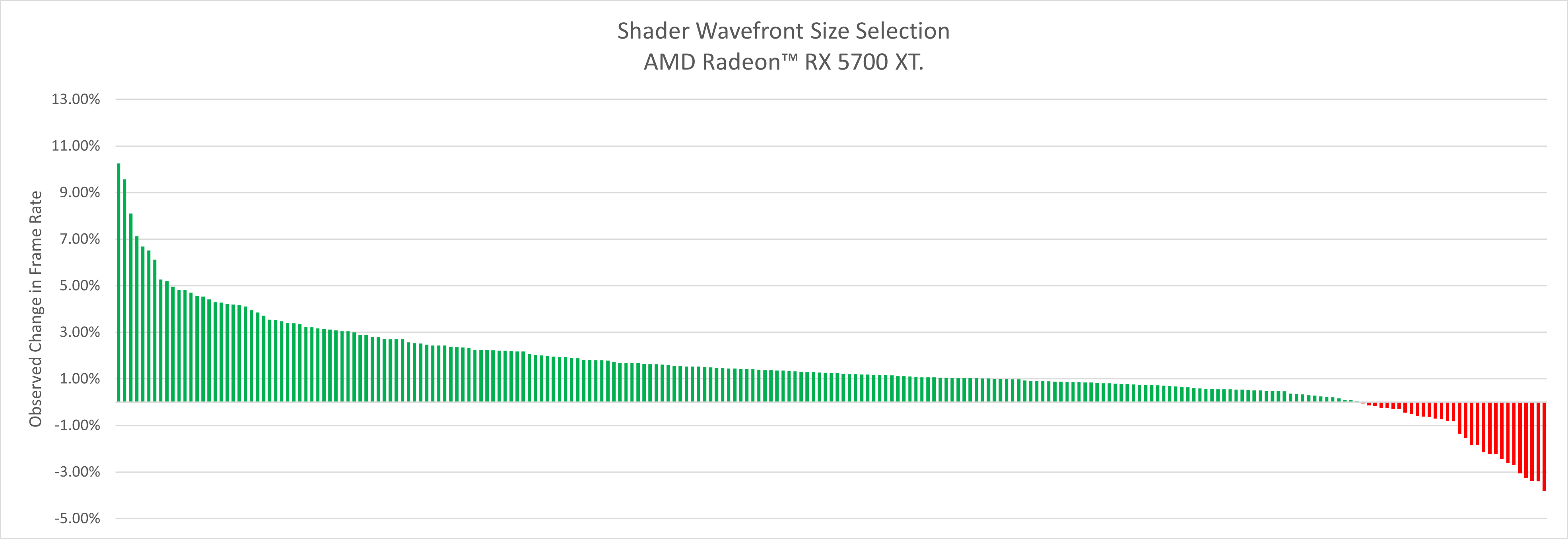}
    \caption{\small{\textbf{Shader Wavefront Size Selection - AMD Radeon\texttrademark~RX 5700 XT.} We provide a histogram of the observed change in frame rate when enabling the RL-based compiler heuristic to select shader wavefront size within the production AMD Radeon\texttrademark~Software graphics compiler across the suite of graphics applications executed on the AMD Radeon\texttrademark~RX 5700 XT.}}
    \label{fig:wavefront_size_selection_5700}
\end{figure*}

\subsection{Stability in Production Compilers}
\label{sec:stability_experiments}

\begin{figure}
    \centering
    \includegraphics[width=0.43\linewidth]{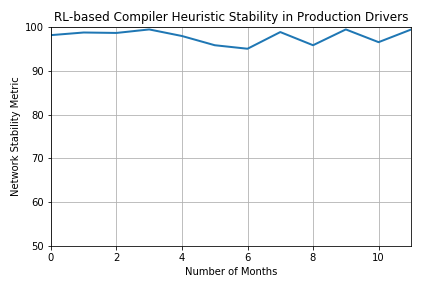}
    \caption{\small{\textbf{Stability in Production Compilers.} We use the inverse of the rate of statistically significant regressions as a metric for the stability of our pre-trained networks when deployed as a heuristic in production compilers, where a value of 100\% indicates no statistically significant regressions. We track this metric over a year of production compiler updates without retraining the deployed model.}}
    \label{fig:stability_metric}
\end{figure}

With the frequent changes to compiler IR, a model needs to be able to generalize to new generated code.
A successfully trained model can be deployed without having to frequently update the learned parameters.
We use the inverse of the rate of statistically significant regressions as a metric for the stability of our pre-trained networks when deployed as a heuristic in production compilers.
We determine statistical significance using a 1-tailed t-test assuming a $p$-value of $5\%$.
Here, 100\% \textit{network stability} indicates that no benchmarks saw a significant regression when using the DNN as the wave size selection heuristic as compared to the default compiler behavior.
Figure~\ref{fig:stability_metric} shows how this stability metric changes over a year of production compiler updates without retraining the deployed model.

\subsection{Generalizing Across Target Hardware}
\label{subsec:experiments-asic}

We use transfer learning to optimize wavefront execution mode selection for another GPU in the AMD Radeon\texttrademark~family - the AMD Radeon\texttrademark~RX 5700 XT.
Rather than starting from a random initialization point, we use the final $Q$-table and accompanying behavior policy ($\pi_\beta$) from the previous experiment as the initialization point.
In only 10 iterations over our set of graphics benchmarks, the learned compiler heuristic matches or surpasses the observed frame rates in {94.4\%} of graphics benchmarks with increases of up to {10.3\%} and an average of {1.5\%}.
{Figure~\ref{fig:wavefront_size_selection_5700}} gives a histogram of the results.
All experiments are run in our AMD production graphics compiler over 270 graphics benchmarks with an average of 230 unique shaders per benchmark.

\section{Discussion}
\label{sec:discussion}

As discussed in Section~\ref{sec:background_compiler_autotuning}, production compilers impose many constraints on ML-based solutions to code optimization problems.
To operate within the limitations of compute and memory resources, we tightly represent the compiler IR of each shader as a feature vector using only 176 bytes and generate RL-based compiler heuristics using a DNN with less than 20KB of learnable parameters.
As shown in Section~\ref{sec:experimental-results}, our RL-based GPU compiler autotuning framework is able to learn stable heuristics that are not only resilient to constantly evolving production software but also generalize well across both graphics benchmarks and GPU targets.

While the average performance uplift on both hardware targets is modest, the results are non-trivial as maximum performance uplift is determined by: (1) the performance of the baseline, (2) the impact of the compiler optimization in question, and (3) the prevalence of the shader under consideration.
In our experiments, we use the fully tuned production driver as a baseline of comparison, which is optimized by teams of experts and often leaves little room for further optimization.
Furthermore, the theoretical maximum speedup is constrained under Amdahl's Law.
As discussed in Section~\ref{sec:background_compiler_autotuning}, modern graphics applications use concurrently executing pipelines of dedicated shaders to render each frame.
Let $p$ denote the percentage of total execution time accounted for by a given pipeline of shaders such that $p \in [0, 1]$.
Let $s$ denote the estimated speedup when applying a set of compiler optimizations to that pipeline of shaders such that $s \geq 0$, where $0 \leq s < 1$ is a performance degradation.
It follows that applying those compiler optimizations to the given pipeline of shaders results in a global speedup of $1 / (1 - p + \frac{p}{s})$ over the entire graphics application.
Due to the constraints of the compiler, we act on each shader in isolation, oblivious to concurrently executing pipelines and even other shaders staged in the same pipeline.
It's often the case in modern graphics applications that individual shaders account for small percentages of the total execution time.
Furthermore, individual compiler optimizations do not impact all shaders equally.
In fact, the back-end compiler is responsible for optimizing the byte code from many different front-end compilers, each of which may have unique or contradicting opportunities for optimization.

As such, we conjecture that the optimal decision function ($\pi^*$) may be intimately tied to the target set of applications ($\mathcal{A}$).
This is contrary to the concept of overfitting, in which a model is too tightly fit to a training set such that it negatively impacts generalization to a validation set.
In such a scenario, the distribution of training and validation data is assumed to be independent and identically distributed (IID).
Because the computational and bandwidth requirements of a graphics application can vary significantly depending on the behavior of shaders and the properties of a scene, we conjecture that shifting $\mathcal{A}$ can lead to drastic shifts in the data distribution that may introduce conflicting opportunities for optimization and may void the IID assumption.
In future work, we aim to explore this relationship further using techniques from out-of-distribution learning for the purpose of uncovering and appropriately handling conflicting opportunities for optimization.

\section{Related Works}
\label{sec:related-works}

Previous works have begun to explore the use of reinforcement learning for compiler optimization~\cite{haj2020neurovectorizer, huang2019autophase,mcgovern2002building,trofin2021mlgo,mendis2019compiler, mirhoseini2017device}.
Huang \textit{et al.}~\cite{huang2019autophase} optimize compiler HLS phase ordering using cycle count reduction as a reward signal to guide learning using a framework they call AutoPhase.
McGovern \textit{et al.}~\cite{mcgovern2002building} use a temporal difference reinforcement learning algorithm to optimize the scheduling of straight-line code.
Mirhoseini \textit{et al.}~\cite{mirhoseini2017device} use on-policy reinforcement learning to optimize device placements for graph partitioning of deep neural networks.
Mendis \textit{et al.}~\cite{mendis2019compiler} use imitation learning to learn how to best convert scalar code into vector code (a process known as vectorization) by mimicking a state-of-the-art solution.
Haj-Ali \textit{et al.}~\cite{haj2020neurovectorizer}, drive the vectorization pass with pragma directives to inform a LLVM compiler heuristic and build an automated feature extraction framework to represent source code using a framework they call NeuroVectorizer.
Trofin \textit{et al.}~\cite{trofin2021mlgo} formulates the inlining-for-size problem as a MDP, using native size reduction as a reward signal to guide an agent to learn the optimal policy for a production LLVM compiler, using a framework they call MLGO.
Aside from~\cite{mendis2019compiler}, which uses imitation learning, each of these previous works use an on-policy reinforcement learning strategy.
While they each yield impressive results on their respective tasks, our solution is an off-policy DRL algorithm based on $Q$-learning that accounts for the unique problems that arise from autotuning production GPU compilers using a time decay with each update step.

\section{Conclusion and Future Work}
\label{sec:conclusion}

We developed and implemented a GPU compiler autotuning framework that uses off-policy deep reinforcement learning to generate heuristics that improve the frame rates of graphics applications.
Our framework combines continuous integration (CI) with $Q$-learning to learn the optimal heuristic settings that maximize the expected frame rate improvements across a set of graphics benchmarks.
By accounting for the the rapid changes in software development, we show that we are able deploy these trained models as stable heuristics in constantly evolving production compilers.
Furthermore, we show that this framework provides generalized performance gains across a large suite of graphics benchmarks across GPUs.
In future work, we aim to explore the relationship between our set of static counters and the dynamic properties the neural network has learned to account for.
Additionally, we aim to extend this framework across domains with continuous action spaces using techniques from deep $Q$-learning.

\section*{Acknowledgements}
We would like to thank Mike Bedy, Robert Gottlieb, Chris Reeve, Andrew Dupont, Karen Dintino, Peter Scannell and the rest of the AMD GPU compiler team for insightful discussions and infrastructure support. \\

\noindent © 2021 Advanced Micro Devices, Inc.  All rights reserved.
AMD, the AMD Arrow logo, Radeon, and combinations thereof are trademarks of Advanced Micro Devices, Inc.
Other product names used in this publication are for identification purposes only and may be trademarks of their respective companies.

\bibliography{citations.bib}

\begin{thebibliography}{10}

\bibitem{bodin1998iterative}
F.~Bodin, T.~Kisuki, P.~Knijnenburg, M.~O'Boyle, and E.~Rohou, ``Iterative
  compilation in a non-linear optimisation space,'' 1998.

\bibitem{leather2020machine}
H.~Leather and C.~Cummins, ``Machine learning in compilers: Past, present and
  future,'' in {\em 2020 Forum for Specification and Design Languages (FDL)},
  pp.~1--8, IEEE, 2020.

\bibitem{stephenson2021cpgo}
M.~Stephenson, R.~Rangan, and S.~Keckler, ``Cooperative profile guided
  optimizations,'' {\em High Performance Graphics}, 2021.

\bibitem{ashouri2018survey}
A.~H. Ashouri, W.~Killian, J.~Cavazos, G.~Palermo, and C.~Silvano, ``A survey
  on compiler autotuning using machine learning,'' {\em ACM Computing Surveys
  (CSUR)}, vol.~51, no.~5, pp.~1--42, 2018.

\bibitem{ashouri2014bayesian}
A.~H. Ashouri, G.~Mariani, G.~Palermo, and C.~Silvano, ``A bayesian network
  approach for compiler auto-tuning for embedded processors,'' in {\em 2014
  IEEE 12th Symposium on Embedded Systems for Real-time Multimedia
  (ESTIMedia)}, pp.~90--97, IEEE, 2014.

\bibitem{bergstra2012machine}
J.~Bergstra, N.~Pinto, and D.~Cox, ``Machine learning for predictive
  auto-tuning with boosted regression trees,'' in {\em 2012 Innovative Parallel
  Computing (InPar)}, pp.~1--9, IEEE, 2012.

\bibitem{cavazos2007rapidly}
J.~Cavazos, G.~Fursin, F.~Agakov, E.~Bonilla, M.~F. O'Boyle, and O.~Temam,
  ``Rapidly selecting good compiler optimizations using performance counters,''
  in {\em International Symposium on Code Generation and Optimization
  (CGO'07)}, pp.~185--197, IEEE, 2007.

\bibitem{cummins2017end}
C.~Cummins, P.~Petoumenos, Z.~Wang, and H.~Leather, ``End-to-end deep learning
  of optimization heuristics,'' in {\em 2017 26th International Conference on
  Parallel Architectures and Compilation Techniques (PACT)}, pp.~219--232,
  IEEE, 2017.

\bibitem{fursin2011milepost}
G.~Fursin, Y.~Kashnikov, A.~W. Memon, Z.~Chamski, O.~Temam, M.~Namolaru,
  E.~Yom-Tov, B.~Mendelson, A.~Zaks, E.~Courtois, {\em et~al.}, ``Milepost gcc:
  Machine learning enabled self-tuning compiler,'' {\em International journal
  of parallel programming}, vol.~39, no.~3, pp.~296--327, 2011.

\bibitem{leather2014automatic}
H.~Leather, E.~Bonilla, and M.~O'boyle, ``Automatic feature generation for
  machine learning--based optimising compilation,'' {\em ACM Transactions on
  Architecture and Code Optimization (TACO)}, vol.~11, no.~1, pp.~1--32, 2014.

\bibitem{monsifrot2002machine}
A.~Monsifrot, F.~Bodin, and R.~Quiniou, ``A machine learning approach to
  automatic production of compiler heuristics,'' in {\em International
  conference on artificial intelligence: methodology, systems, and
  applications}, pp.~41--50, Springer, 2002.

\bibitem{song2020learning}
H.~Song, M.~Kim, D.~Park, and J.-G. Lee, ``Learning from noisy labels with deep
  neural networks: A survey,'' {\em arXiv preprint arXiv:2007.08199}, 2020.

\bibitem{zhang2016understanding}
C.~Zhang, S.~Bengio, M.~Hardt, B.~Recht, and O.~Vinyals, ``Understanding deep
  learning requires rethinking generalization,'' {\em arXiv preprint
  arXiv:1611.03530}, 2016.

\bibitem{zhu2004class}
X.~Zhu and X.~Wu, ``Class noise vs. attribute noise: A quantitative study,''
  {\em Artificial intelligence review}, vol.~22, no.~3, pp.~177--210, 2004.

\bibitem{graesser2019foundations}
L.~Graesser and W.~L. Keng, {\em Foundations of Deep Reinforcement Learning:
  Theory and Practice in Python}.
\newblock Addison-Wesley Professional, 2019.

\bibitem{lin2020deep}
E.~Lin, Q.~Chen, and X.~Qi, ``Deep reinforcement learning for imbalanced
  classification,'' {\em Applied Intelligence}, pp.~1--15, 2020.

\bibitem{zhao2016deep}
D.~Zhao, Y.~Chen, and L.~Lv, ``Deep reinforcement learning with visual
  attention for vehicle classification,'' {\em IEEE Transactions on Cognitive
  and Developmental Systems}, vol.~9, no.~4, pp.~356--367, 2016.

\bibitem{haj2020neurovectorizer}
A.~Haj-Ali, N.~K. Ahmed, T.~Willke, Y.~S. Shao, K.~Asanovic, and I.~Stoica,
  ``Neurovectorizer: end-to-end vectorization with deep reinforcement
  learning,'' in {\em Proceedings of the 18th ACM/IEEE International Symposium
  on Code Generation and Optimization}, pp.~242--255, 2020.

\bibitem{huang2019autophase}
Q.~Huang, A.~Haj-Ali, W.~Moses, J.~Xiang, I.~Stoica, K.~Asanovic, and
  J.~Wawrzynek, ``Autophase: Compiler phase-ordering for hls with deep
  reinforcement learning,'' in {\em 2019 IEEE 27th Annual International
  Symposium on Field-Programmable Custom Computing Machines (FCCM)},
  pp.~308--308, IEEE, 2019.

\bibitem{mcgovern2002building}
A.~McGovern, E.~Moss, and A.~G. Barto, ``Building a basic block instruction
  scheduler with reinforcement learning and rollouts,'' {\em Machine learning},
  vol.~49, no.~2-3, pp.~141--160, 2002.

\bibitem{trofin2021mlgo}
M.~Trofin, Y.~Qian, E.~Brevdo, Z.~Lin, K.~Choromanski, and D.~Li, ``Mlgo: a
  machine learning guided compiler optimizations framework,'' {\em arXiv
  preprint arXiv:2101.04808}, 2021.

\bibitem{watkins1992q}
C.~J. Watkins and P.~Dayan, ``Q-learning,'' {\em Machine learning}, vol.~8,
  no.~3-4, pp.~279--292, 1992.

\bibitem{mnih2013playing}
V.~Mnih, K.~Kavukcuoglu, D.~Silver, A.~Graves, I.~Antonoglou, D.~Wierstra, and
  M.~Riedmiller, ``Playing atari with deep reinforcement learning,'' {\em arXiv
  preprint arXiv:1312.5602}, 2013.

\bibitem{kumar2020conservative}
A.~Kumar, A.~Zhou, G.~Tucker, and S.~Levine, ``Conservative q-learning for
  offline reinforcement learning,'' {\em arXiv preprint arXiv:2006.04779},
  2020.

\bibitem{gleeson2021rl}
J.~Gleeson, M.~Gabel, G.~Pekhimenko, E.~de~Lara, S.~Krishnan, and
  V.~Janapa~Reddi, ``Rl-scope: Cross-stack profiling for deep reinforcement
  learning workloads,'' {\em Proceedings of Machine Learning and Systems},
  vol.~3, 2021.

\bibitem{kumar2020discor}
A.~Kumar, A.~Gupta, and S.~Levine, ``Discor: Corrective feedback in
  reinforcement learning via distribution correction,'' {\em arXiv preprint
  arXiv:2003.07305}, 2020.

\bibitem{alon2019code2vec}
U.~Alon, M.~Zilberstein, O.~Levy, and E.~Yahav, ``code2vec: Learning
  distributed representations of code,'' {\em Proceedings of the ACM on
  Programming Languages}, vol.~3, no.~POPL, pp.~1--29, 2019.

\bibitem{ben2018neural}
T.~Ben-Nun, A.~S. Jakobovits, and T.~Hoefler, ``Neural code comprehension: A
  learnable representation of code semantics,'' {\em arXiv preprint
  arXiv:1806.07336}, 2018.

\bibitem{cummins2020programl}
C.~Cummins, Z.~V. Fisches, T.~Ben-Nun, T.~Hoefler, and H.~Leather, ``Programl:
  Graph-based deep learning for program optimization and analysis,'' {\em arXiv
  preprint arXiv:2003.10536}, 2020.

\bibitem{venkatakeerthy2020ir2vec}
S.~VenkataKeerthy, R.~Aggarwal, S.~Jain, M.~S. Desarkar, R.~Upadrasta, and
  Y.~Srikant, ``Ir2vec: Llvm ir based scalable program embeddings,'' {\em ACM
  Transactions on Architecture and Code Optimization (TACO)}, vol.~17, no.~4,
  pp.~1--27, 2020.

\bibitem{mendis2019compiler}
C.~Mendis, C.~Yang, Y.~Pu, S.~Amarasinghe, and M.~Carbin, ``Compiler
  auto-vectorization with imitation learning,'' 2019.

\bibitem{mirhoseini2017device}
A.~Mirhoseini, H.~Pham, Q.~V. Le, B.~Steiner, R.~Larsen, Y.~Zhou, N.~Kumar,
  M.~Norouzi, S.~Bengio, and J.~Dean, ``Device placement optimization with
  reinforcement learning,'' in {\em International Conference on Machine
  Learning}, pp.~2430--2439, PMLR, 2017.

\end{thebibliography}

\end{document}